\documentclass{IEEEtran}
\usepackage{url}
\usepackage{graphicx}
\usepackage{subfigure}
\usepackage{multirow}
\begin{document}

\title{Beyond Bags of Words: Inferring Systemic Nets}
\author{\IEEEauthorblockN{D.B. Skillicorn}
	\IEEEauthorblockA{School of Computing \\
		Queen's University \\
	Kingston.  Canada}
\and
\IEEEauthorblockN{N. Alsadhan}
	\IEEEauthorblockA{School of Computing \\
		Queen's University \\
	Kingston.  Canada}
}

\maketitle

\begin{abstract}
Textual analytics based on representations of documents as bags of words
have been reasonably successful.
However, analysis that requires deeper insight into language, into
author properties, or into the contexts in which documents were 
created requires a richer representation.
Systemic nets are one such representation.
They have not been extensively used because they required
human effort to construct.
We show that systemic nets can be algorithmically inferred from corpora, 
that the resulting nets are plausible, and that they can provide practical
benefits for knowledge discovery problems.
This opens up a new class of practical analysis techniques for
textual analytics.
\end{abstract}

\section{Motivation}

Sets of documents represent one of the largest sources of ``big data",
with web search engines indexing tens of billions of pages.
For information retrieval, where the key problem is to identify
the content of a document, the bag-of-words model of text has proven
extremely successful, even for languages such as English where word order
is crucial to meaning.
Sentences such as ``the criminal shot the officer" and 
``the officer shot the criminal" are equally plausible
responses to queries about criminals and officers, but
much less equivalent from the perspective of, say, the media.

Many textual analytics tasks assess documents based not 
just on what they contain,
but how they were built -- their significance depends on properties
that derive from the author, the author's goals or intent, and the
situation or context in which the document was created.
Examples include: determining a document's authorship, an author's
gender or age, an author's opinions (polarity, sentiment),
an author's intention, and whether the author is being knowingly
deceptive. Such tasks are key to domains such as
e-discovery (finding significant emails in a corporate archive),
intelligence (finding meaningful threats in a set of forum posts), measuring
the effectiveness of a marketing campaign (in online social media
posts), or predicting an uprising (using Twitter feed data).

Although bag-of-words approaches have been moderately successful
for such problems, they tend to hit a performance wall (80\%
prediction accuracy is typical) because the representation fails
to capture sufficient subtleties \cite{tausczik}.
There have been attempts to increase the quality of representations, 
for example by extracting parse trees (that is, context-free grammar 
representations) but this focuses entirely on (somewhat artificial) 
language structure, and not at all on mental processes \cite{kanayama}. 
Other approaches leverage syntactically
expressed semantic information, for example 
by counting word bigrams, 
by using Wordnet \cite{scott:matwin}, or using deep learning
\cite{socher}.

One approach that shows considerable promise is
\emph{systemic functional linguistics} 
\cite{sfl,halliday:webster,davies:sfl},
a model of language generation with sociological origins and
an explicit focus on the effect of the creator's mental state and social
setting on a created document.
In this model, the process of generating an utterance (a sentence,
a paragraph, or an entire document) is conceived of as traversing
a \emph{systemic net}, a set of structured choices.
The totality of these choices defines the created utterance.
At some nodes, the choice is disjunctive: continue by choosing
\emph{this} option or by choosing \emph{that} one.
At others, the choice is conjunctive: choose a subset
of these options and continue in parallel down several paths.

\begin{figure}
\begin{center}
\includegraphics[width=0.4\textwidth]{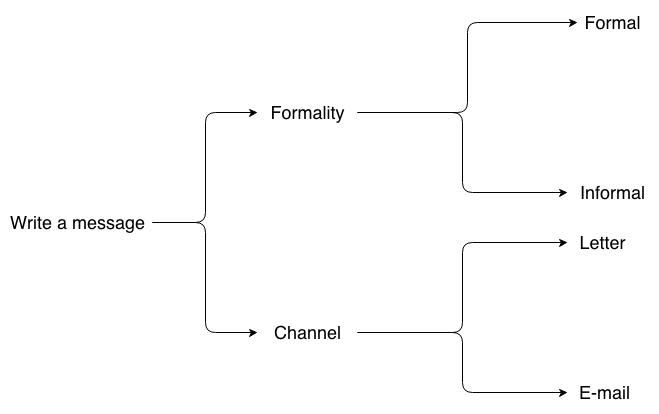}
\end{center}
\caption{A simple example of a systemic net}
\label{simpleexample}
\end{figure}

Figure~\ref{simpleexample} shows a simple example of a systemic net.
The decision to communicate requires a parallel (independent) choice of the
level of formality to be used and the communication
channel to be used. Each of these choices requires a further choice.
The level of formality could be formal or informal; and the channel
could be to use a letter or an email. These choices at the second
level are disjunctive -- it has to be one or the other. Further choices
exist below these ones, so the systemic net notionally continues to
the right until it results in concrete language.

Production using a context-free grammar also requires a structured
set of choices, but the choices are top-down (so that the first
choice is to instantiate, for example, a declarative sentence as a subject,
an object, and a verb). In contrast, the order of the choices in a systemic
net has no necessary relationship to the concreteness of the implications
of those choices.
For example, the choice to use formal or informal style is
an early choice with broad consequences that limit the possibilities
for subsequent choices.
The choice to write a letter or an email
is also an early choice but its immediate consequence
is narrow and low level: typically whether the first word of the
resulting document will be ``Dear" (for a letter) or not
(for an email).

\begin{figure}
\begin{center}
\includegraphics[width=0.5\textwidth]{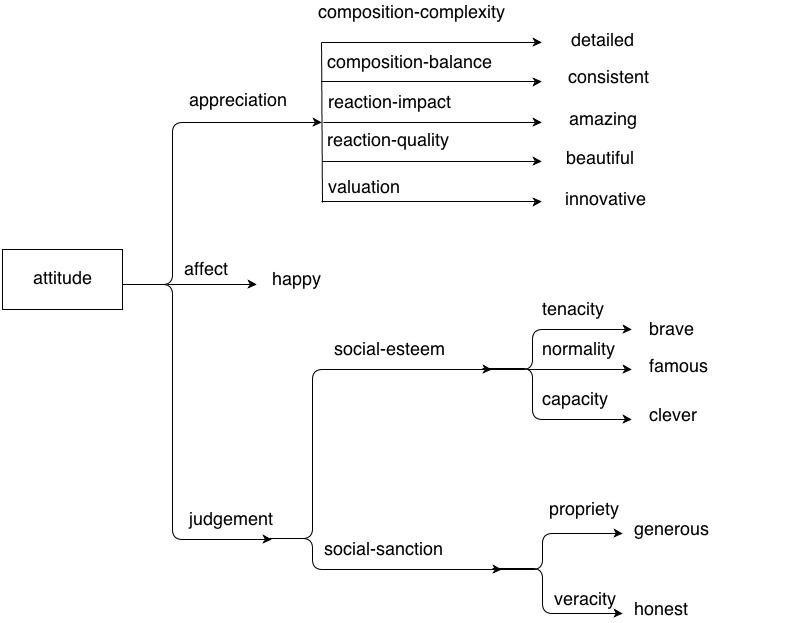}
\end{center}
\caption{The Appraisal systemic net, appropriate for reviews}
\label{appraisal}
\end{figure}

An example of a well-used systemic net, called the
Appraisal Net \cite{lexicalpredictors},
is shown in Figure~\ref{appraisal}. It describes the way in which
choices of adjectives are made when evaluating some object.
The choice process is not arbitrary; rather an individual
chooses simultaneously from up to three parallel paths: 
appreciation, affect, and judgement.
Within two of these choices, there are
then subsequent parallel choices that lead to particular adjectives
-- one example adjective is shown at each leaf.
These choices are associated with different aspects of the situation:
composition-complexity captures aspects of the object being appraised,
while reaction-quality captures aspects of the person doing the
appraising.

The power of systemic nets comes because these choices are made,
not simply for the goal of constructing a syntactically valid
sentence, but because of the limitations and exigencies of
\emph{social purpose} (certain things cannot be said in certain
circumstances although syntactically valid); \emph{mental
state} (because language generation is a largely subconscious
process), and the properties of the language in use.
In other words, the choice of adjective in an appraisal
certainly says something about the object being assessed, 
but also reveals something about the person
doing the assessing; and the structure of the choices would be
different in English and, say, French or Japanese.

A systemic net is explanatory at three different levels.
First, the existence of a net organizes constructions into
categories and so explains some aspects of how the pieces in
a text fit together.

Second, the choices made by individuals traversing a net are not
typically unique; rather, they cluster into common choice
patterns that reflect particular kinds of textual targets.
This is because there are social rules that govern acceptable
end-products. Each individual can write with an individual style,
but we can also say that some set of documents by different
authors are written in a down-to-earth style, and another
set in a flowery style. This idea of a consistent
set of choices in a net, leading to detectable consistencies in
the resulting documents is called a \emph{register}.
Thus the set of registers associated with a net are also explanatory.

Third, for any particular document we can list the choices made
in its construction, and this becomes a record that describes
that document at a higher level of abstraction than as a bag of words.
This level of explanation is most directly useful for knowledge
discovery -- such a description of choices can be used for clustering or for
prediction.

Thus the advantages of a systemic functional approach to textual
analytics are:
\begin{itemize}
\item
The choices within the net are a smaller, more abstract, and more 
structured set than the choice of individual words, 
and therefore provide a stronger
foundation for knowledge discovery -- a kind of structured attribute
selection; and
\item
These choices reflect, and make accessible, the mental state of the
author/speaker and his/her perception of the social situation for which
the text was constructed. This enables a kind of reverse engineering
of how the text came to be, that is knowledge discovery about
authors and settings.
\end{itemize}

The reason why systemic net approaches have not been more widely
used in textual analytics is primarily because they have, so far,
been constructed by computational linguists, often requiring
several person-years to build, even when modestly sized.
Some substantial systemic nets have been built, but usually within
the context of projects where they have been kept confidential;
those that are public, like the Appraisal Net above, are usually
small.

The contribution of this paper is twofold:
\begin{itemize}
\item
We show that it is possible to infer systemic nets from corpora
using Non-Negative Matrix Factorization (NNMF), and that these
nets are plausible.
Thus we are able to construct systemic nets for any corpus, and for
any set of relevant words. This creates a new path to representing
corpora at a deeper level, but without the need (and cost) for substantial
human input.
\item
We show that the resulting systemic nets organize corpora
more strongly than the corresponding bags of words, and that
this organization improves both clustering and prediction tasks,
using authorship prediction as a demonstration task.
\end{itemize}
This removes the bottleneck to the widespread use of systemic-network
approaches, which opens up large, variable corpora to new, deeper
levels of analysis.

Section 2 describes some of the related work leveraging systemic
nets for knowledge discovery tasks.
In Section 3 we introduce the process for inducing systemic nets
from data and show the results on a modest corpus of novels.
In Section 4, we demonstrate some of the benefits of using the resulting
nets.

\section{Related work}

There have been several applications of predefined systemic nets
to textual prediction problems. For example, Whitelaw \emph{et al.}
\cite{whitelaw}
show improvement in sentiment analysis using the Appraisal Net
mentioned above.
Argamon \emph{et al.} show how to predict personality type from
authored text, again using systemic functional ideas
\cite{whitelaw:argamon}.
Herke-Couchman and Patrick derive interpersonal distance from
systemic network attributes \cite{herke}.

The most successful application of systemic functional techniques
is the Scamseek project. The goal of this project was to predict,
with high reliability, web pages that represented financial scams
and those that represented legitimate financial products. This is a
challenging problem -- the differences between the two classes are
small and subtle, and even humans perform poorly at the margins.
The fraction of documents representing scams was less than 2\% of the
whole.
This project's predictive model was successfully deployed on behalf of the
Australian Securities and Investments Commission \cite{scamseek}.
However, the effort to construct the registers corresponding to
normal and (many varieties of) scam documents was substantial.

Kappagoda \cite{kappagoda} shows that word-function tags can be added
to words using conditional random fields, in the same kind of general
way that parsers add part-of-speech tags to words. These word-function
tags provide hints of the systemic-functional role that words carry.
This is limited because there is no hierarchy. Nevertheless, he is able
to show that the process of labelling can be partially automated
and that the resulting tags aid in understanding documents.

\section{Inductive discovery of systemic nets}

The set of choices in a systemic net lead eventually, at the leaves,
to choices of particular (sets of) words. One way to conceptualize
a systemic net, therefore, is as a hierarchical clustering of words,
with each choice representing selection of a subset\footnote{Complete
systemic nets also include a downstream phase that defines
the process for assembling the parts of a constructed document
into its actual linear sequence. We ignore this aspect. In
declarative writing, assembly is usually straightforward,
although this is not the case in, for example, poetry.}. We use this
intuition as a way to inductively construct a systemic net: words that
are used together in the same document (or smaller unit such as a sentence
or paragraph) are there because of a particular sequence of choices.
An inductive, hierarchical clustering can approximate
a hierarchical set of choices.

Our overall strategy is to build document-word matrices
(where the document may be as small as a single sentence), and then
cluster the columns (that is, the words) of such matrices using
the similarity of the documents in which they appear. The question
then is: which clustering algorithm(s) to use.

In this domain, similarity between a pair of documents depends
much more strongly on the \emph{presence} of words
than on their \emph{absence}. Conventional clustering
algorithms, for example agglomerative hierarchical clustering and
other algorithms that use distance as a surrogate for similarity,
are therefore not appropriate, since mutual absence of a word
in two different documents is uninformative, but increases their
apparent similarity.

Singular value decomposition is reasonably effective
(J.L. Creasor, unpublished work) but there are major issues 
raised by the need to normalize the document-word
matrix so that the cloud of points it represents is centered around
the origin. Typical normalizations such as z-scoring conflate
median frequencies with zero frequencies and so introduce artifacts
that are difficult to compensate for in subsequent analysis.

We therefore chose to use Non-Negative Matrix Factorization, since a
document-word matrix naturally has non-negative entries.
An NNMF decomposes a document-word matrix, $A$, as the product
of two other matrices:
\[
A \;=\; W H
\]
If $A$ is $n \times m$, then $W$ is $n \times r$ for some chosen
$r$ usually much smaller than either $m$ or $n$, and $H$ is
$r \times m$. All of the entries of $W$ and $H$ are non-negative,
and there is a natural interpretation of the rows of $H$ as
`parts' that are `mixed' together by each row of $W$ to give the
observed rows of $A$ \cite{lee:seung:nature}.

Algorithms for computing an NNMF are iterative in nature, and
the results may vary from execution to execution because of the
random initialization of the values of $W$ and $H$. In general,
the results reported here are obtained by computing the NNMF
10 times and taking the majority configuration.
We use a conjugate gradient version of NNMF, using Matlab code
written by Pauca and Plemmons.

There are two alternative ways to use an NNMF, either directly
from the given data matrix, or starting from its transpose.
If we compute the NNMF of the transpose of A, we obtain:
\[
A' \;=\; \bar{W} \bar{H}
\]
and, in general, it is not the case that $\bar{H} = W'$
and $\bar{W} = H'$.
Experiments showed that results were consistently better if
we applied the NNMF to $A'$, that is to the word-document matrix.
The textual unit we use is the paragraph. A single sentence
might, in some contexts, be too small; a whole document is
too large since it reflects thousands of choices.

We extracted paragraph-word matrices in two ways. A
parts-of-speech-aware tagger made it possible to extract the
frequencies of, for example, all pronouns or all determiners
\cite{qtagger}.
For larger word classes, such as adjectives, it was also possible
to provide the tagger with a given list and have it extract only
frequencies of the provided words.
Frequency entries in each matrix were normalized by the total
number of words occurring in each paragraph, turning word counts
into word rates. This compensates
for the different lengths of different paragraphs.

Superior results were obtained
by choosing only $r = 2$ components.
In the first step,
the $\bar{W}$ matrix has dimensionality \emph{number of words} $\times \; 2$,
with non-negative entries.
Each word was allocated to the cluster (column) with the largest entry in the
corresponding row of $\bar{W}$, and the process repeated with the two
submatrices obtained by splitting the rows of $A'$ based on this
cluster allocation.
This process continued until the resulting clusters could not
be cleanly separated further.
These clusters therefore form a binary tree where each internal
node contains the union of the words of its two children.

Each NNMF was repeated 10 times to account for
the heuristic property of the algorithm. We were able to leverage
this to estimate the confidence of each clustering.
For example, there were occasionally particular words whose membership
oscillated between two otherwise stable clusters, and this provided a
signal that they didn't fit well with either.
We were also able to use this to detect when to stop the recursive
clustering: either clusters shrank until they contained only a
single word (usually a high-frequency one), or their subclusters
began to show no consistency between runs, which we
interpreted to mean that the cluster was being over-decomposed.

The result of applying this recursive NNMF algorithm to a
word-paragraph matrix is a hierarchical binary tree whose internal
nodes are interpreted as choice points, and whose leaves represent the
``outputs" that result from making the choices that result
in reaching that leaf.
A leaf consists of a set of words that are considered to be, in a
sense, equivalent or interchangeable from the point of view of
the total set of words being considered.
However, this view of leaves contains a subtle point. Suppose that
a leaf contains the words ``red" and ``green". These are clearly
not equivalent in an obvious sense, and in any given paragraph it is
likely that an author will select only one of them.
In what sense, then, are they equivalent? The answer is that,
from the author's point of view, the choice between them is a trivial
one: either could serve in the context of the document (fragment)
being created. Thus a leaf in the systemic net contains a set of words
from which sometimes a single word is chosen and sometimes a number
of words are chosen -- but in both cases the choice is unconstrained
by the setting (or at least undetectably unconstrained in the
available example data).

We have remarked that choices at internal nodes in a systemic net
can be disjunctive or conjunctive. However, in our construction method
each word in a particular document is allocated to exactly one
cluster or the other.
We estimate the extent to which a choice point is conjunctive or
disjunctive by counting how often the choice goes either way
across the entire set of documents, that is we treat conjunction/disjunction
as a global, rather than a local, property.
(It would be possible to allocate a word to both clusters if
the entries in the corresponding row of $\bar{W}$ had
similar magnitude, and therefore detect conjunctive choices
directly. However, deciding what constitutes a similar magnitude
is problematic because of the variation between runs deriving from
the heuristic nature of the algorithm.)

\section{Inferred systemic nets}

The data used for proof of concept of this approach is a set of 17
novels downloaded from
\url{gutenberg.org} and lightly edited to remove site-specific
content. These novels covered a period of about a century from the
1830s to the 1920s and represent well-written, substantial documents.
For processing they were divided into paragraphs; because of the
prevalence of dialogue in novels, many of these paragraphs are
actually single sentences of reported speech. The total number
of paragraphs is 48,511. The longest novel contained 13,617
paragraphs (\emph{Les Miserables}) and the shortest 736 (\emph{The
39 Steps}).

We selected six different categories of words for experiments
as shown in Table~\ref{wordsets}.

\begin{table}
\begin{center}
  \caption{List of words used to create the systemic networks}
  \label{wordsets}
   \begin{tabular}{| l | p{5cm} |}
      \hline
      Group type & Words \\ \hline
      Personal Pronouns & I, me, my, mine, myself, we, us, our, ours, ourselves, you, your, yours, yourself, yourselves, they, their, theirs, them, themselves, he, him, his, himself, she, her, hers, herself, it, its, itself, one, one's \\  \hline
      Adverbs & afterwards, already, always, immediately, last, now, soon, then, yesterday, above, below, here, outside, there, under, again, almost, ever, frequently, generally, hardly, nearly, never, occasionally, often, rarely \\ \hline
      Auxiliary verbs & was, wasn't, had, were, hadn't, did, didn't, been, weren't, are, is, does, am, has, don't, haven't, doesn't, aren't, do, isn't, have, be,  hasn't \\ \hline
      Positive auxiliary verbs & was, had, were, did, been, is, does, are, am, has, do, have, be \\ \hline
      Adjectives & good, old, little, own, great, young, long, such, dear, poor, new, whole, sure, black, small, full, certain, white, right, possible, large, fresh, sorry, easy, quite, blue, sweet, late, pale, pretty \\ \hline
      Verbs & said, know, see, think, say, go, came, make, come, went, seemed, made, take, looked, thought, saw, tell, took, let, going, get, felt, seen, give, knew, look, done, turned, like, asked \\ \hline
      \end{tabular}
  \end{center}
  \end{table}

\begin{figure}
\begin{center}
\includegraphics[width=0.5\textwidth]{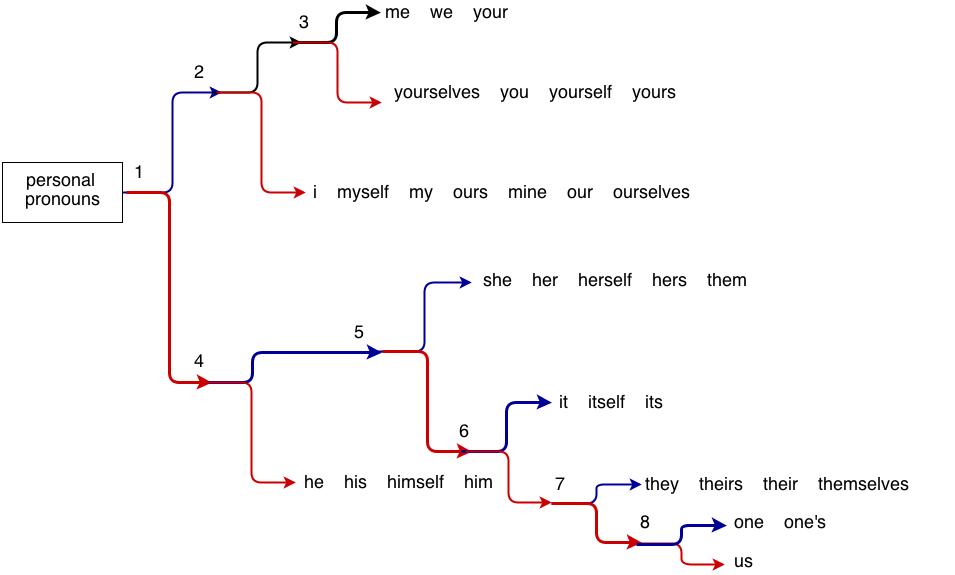}
\end{center}
\caption{Systemic net inferred for pronouns}
\label{pronouns}
\end{figure}

Figure~\ref{pronouns} shows the systemic net of pronouns. In all of these
figures, the thickness of each line indicates how often the corresponding
path was taken as the result of a choice. Lines in blue represent
the ``upper" choice, red the ``lower" choice, and black the situation
where both choices occurred with approximately equal frequency.

The top-level choice (1) in this net is between pronouns
where the point of view is internal to the story,
and where the point of view is of an external narrator. This seems
plausible, especially in the context of novels.
Choice point 2 is largely between first-person and second-person 
pronouns, with apparently anomalous placement of ``me" and``we".
Choice point 4 is between masculine pronouns and others, again
entirely plausible given the preponderance of masculine protagonists
in novels of this period. The remaining choices in this branch separate
feminine, impersonal, and third-person plural pronouns.
All of these choices are strongly disjunctive, weakening down the
tree with choice point 7 the least disjunctive. It might be expected
that, after the choice at point 1, choices might become more conjunctive
as two or more people are mentioned. However, reported speech by one
person is the most common paragraph structure in these novels, and many
of these do not contain another pronoun reference
(``He said `What's for dinner?'").

\begin{figure}
\begin{center}
\includegraphics[width=0.5\textwidth]{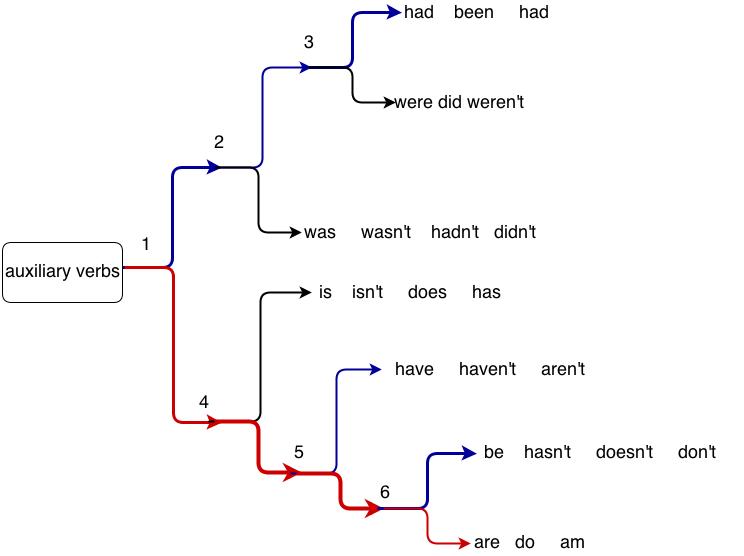}
\end{center}
\caption{Systemic net inferred for auxiliary verbs}
\label{aux}
\end{figure}

\begin{figure}
\begin{center}
\includegraphics[width=0.4\textwidth]{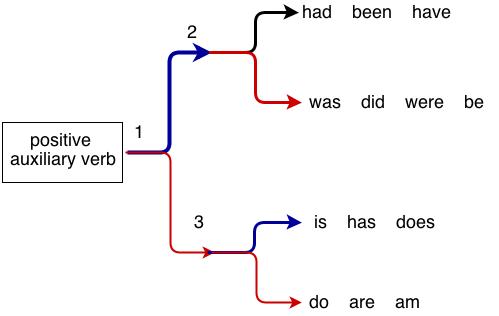}
\end{center}
\caption{Systemic net inferred for positive auxiliary verbs}
\label{auxpos}
\end{figure}

Figure~\ref{aux} shows the systemic net for auxiliary verbs.
These might have separated based on their root verb (to be, to have,
to do) \footnote{and a computational linguist might have
chosen this separation as the most `natural'}
but in fact they separate based on tense. Choice point 1 is
between past tense forms and present tense forms.
Choices between verb forms are visible at the subsequent levels.
Of course, auxiliary verbs are difficult to categorize because
they occur both as auxiliaries, and as stand-alone verbs.

The set of auxiliary verbs is also difficult because many of them
encapsulate a negative (``hadn't"), 
and negatives represent an orthogonal category
of choices. Figure~\ref{auxpos} shows that systemic net when only
the positive auxiliary verbs are considered. Again, tense is the
dominant choice.

\begin{figure}
\begin{center}
\includegraphics[width=0.5\textwidth]{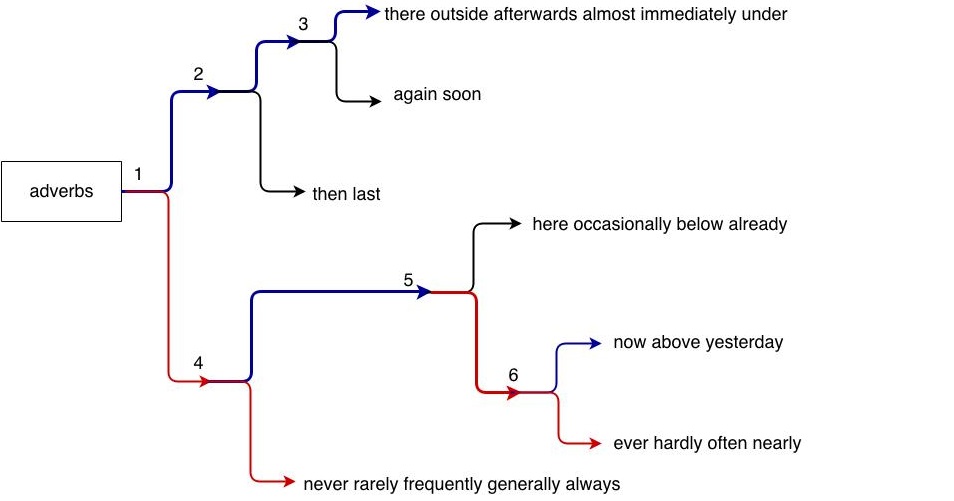}
\end{center}
\caption{Systemic net inferred for adverbs}
\label{adverbs}
\end{figure}

Figure~\ref{adverbs} shows the systemic net for adverbs from a
limited set of three different kinds: time, place, and frequency.
This systemic net seems unclear, but note that at least some branches
agree with intuition, for example the lower branch from choice 4.

\begin{figure}
\begin{center}
\includegraphics[width=0.49\textwidth]{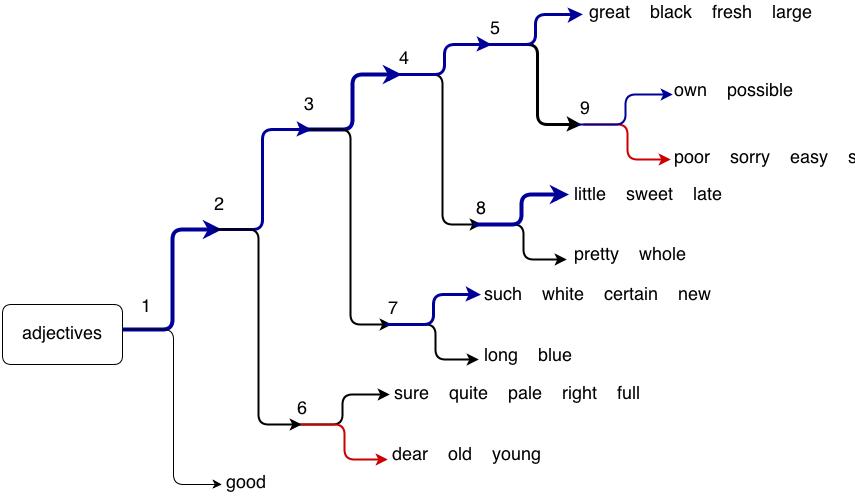}
\end{center}
\caption{Systemic net inferred for adjectives}
\label{adjectives}
\end{figure}

There are a very large number of adjectives used in the corpus, most of
them only rarely.
However, it is interesting to consider how adjectives might be
empirically distinguished in fiction. (Note that this would not
be the same net as the Appraisal Net described earlier, which might
be inferrable from, say, a corpus of product reviews.)
Figure~\ref{adjectives} shows the systemic net for a limited
set of adjectives of three kinds: appearance, color, and time.
This net shows the typical structure for an extremely common word, in this
case ``good" which appears as one outcome of the first choice.
The sets of adjectives at each leaf are not those that would be
conventionally grouped, but there are a number of interesting
associations: ``great" and ``large" occur together, but co-occur with
``black" which is a plausible psychological association.

\begin{figure}
\begin{center}
\includegraphics[width=0.5\textwidth]{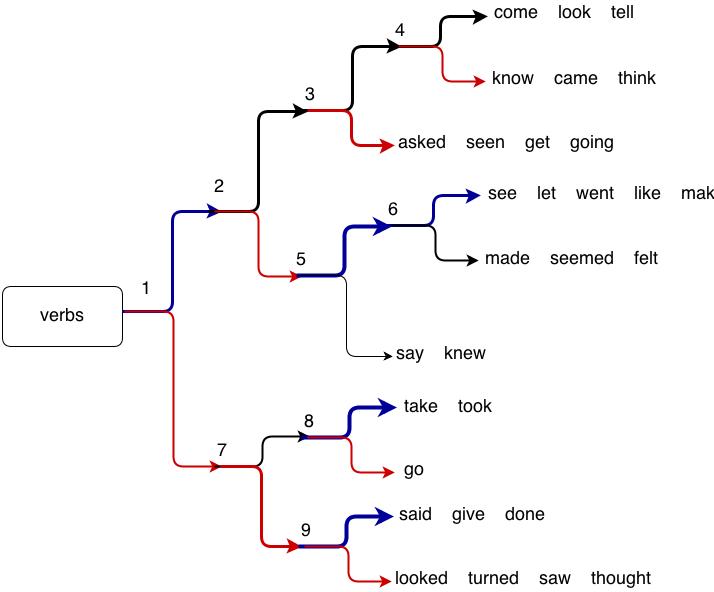}
\end{center}
\caption{Systemic net inferred for verbs}
\label{verbs}
\end{figure}

These systemic nets look, from a human perspective, somewhere between
plausible and peculiar. We now turn to more rigorous validation.
Our goal is not so much that these nets should be explanatory from
an intuitive perspective, but
that they should be useful for knowledge-discovery tasks.

\section{Validation}

To validate our technique for inferring systemic nets, we use
the following methods:
\begin{itemize}
\item
Face validation. The systemic nets should involve choices
that appear sensible and realistic. Note that this does not mean
that they should match the hierarchy created to explain English
grammar -- such a grammar is an artificial construct intended to
suggest consistent rules, and owing much to the grammar of Latin,
rather than an accurate description of how English actually works.
\item
Comparison of document clustering based on word choices and based
on systemic net choices. If choices reflect deeper structure, then
documents should cluster more strongly based on choice structure than
on word structure.
\item
Comparison of the performance of an example prediction
task, authorship prediction, using word
choices and systemic net choices. If choices reflect deeper structure,
it should be easier to make predictions about documents based on
choice structure than on word structure.
\item
Comparison with randomly created choice nets.
Hierarchical clusterings with the same macroscopic structure as
induced systemic nets should perform worse than the induced systemic
nets.
\end{itemize}

\subsection{Face validation}

The systemic nets shown in the previous section are not necessarily what
a linguist might have expected, but it is clear that they capture
regularities in the way words are used (especially in the domain of
novels that was used, with their emphasis on individuals and their
high rates of reported speech).

\subsection{Clustering using word choices versus net choices}

The difference between the systemic net approach and the bag-of-words
approach is that they assume a different set of choices that led to
the words that appear in each paragraph. The bag-of-words model
implicitly assumes that each word was chosen independently; the
systemic net model assumes that each word was chosen based on
hierarchical choices driven by purpose, social setting, mental state,
and language possibilities. Clustering paragraphs based on these
two approaches should lead to different clusters, but those derived
from systemic net choices should be more obvious. In particular, choices
are not independent both because of hierarchy and because of the
extrinsic constraints of the setting (novels, in this case) -- so we
expect to see clusters corresponding to registers.

We used two novels for testing purposes: \emph{Robinson Crusoe} and
\emph{Wuthering Heights}, processed in the same way as our training
data. Since these novels were not used to infer the systemic nets,
results obtained using them show that the nets are capturing some
underlying reality of this document class.

\begin{figure}
\begin{center}
\includegraphics[width=0.45\textwidth]{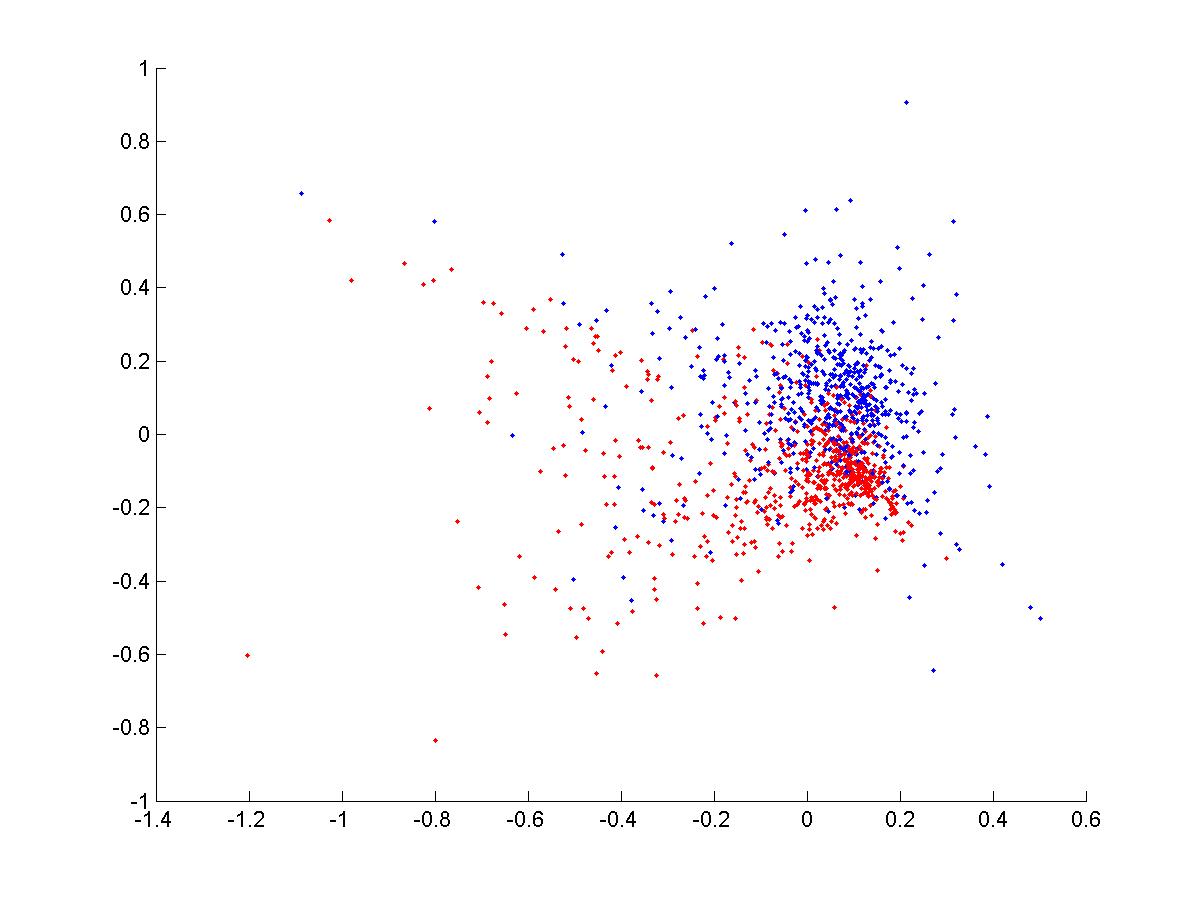}
\includegraphics[width=0.45\textwidth]{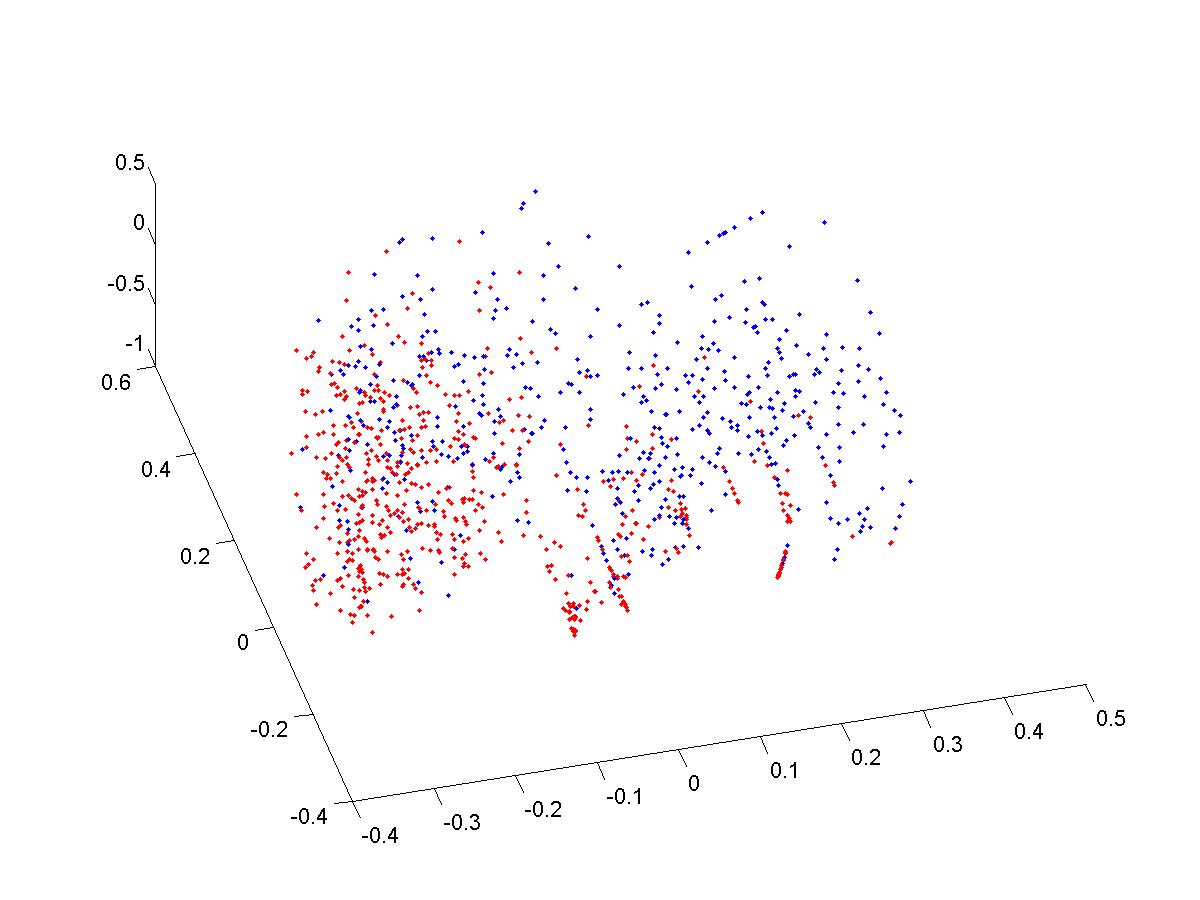}
\end{center}
\caption{SVD using pronouns, bag-of-words above, choices below}
\label{svdpronouns}
\end{figure}

\begin{figure}
\begin{center}
\includegraphics[width=0.45\textwidth]{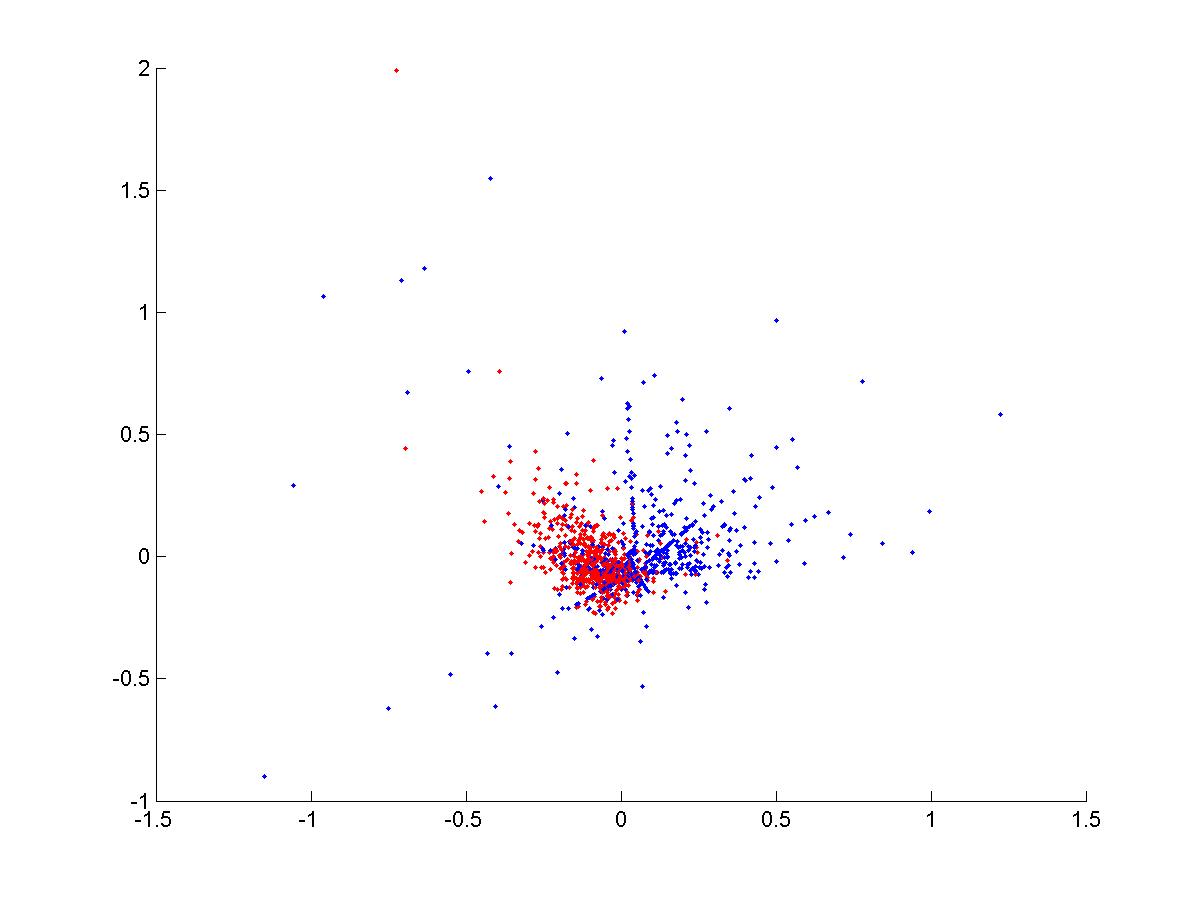}
\includegraphics[width=0.45\textwidth]{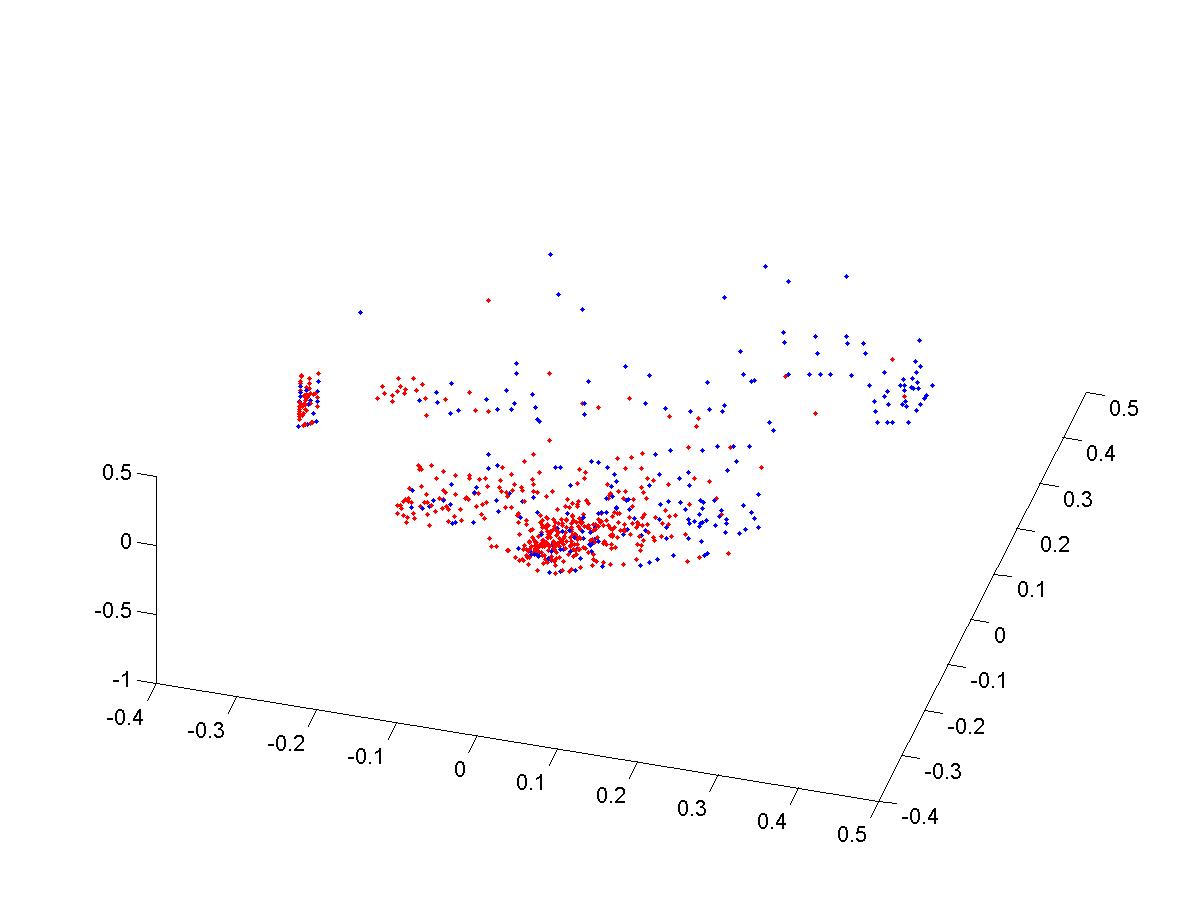}
\end{center}
\caption{SVD using auxiliary verbs, bag-of-words above, choices below}
\label{svdaux}
\end{figure}

\begin{figure}
\begin{center}
\includegraphics[width=0.45\textwidth]{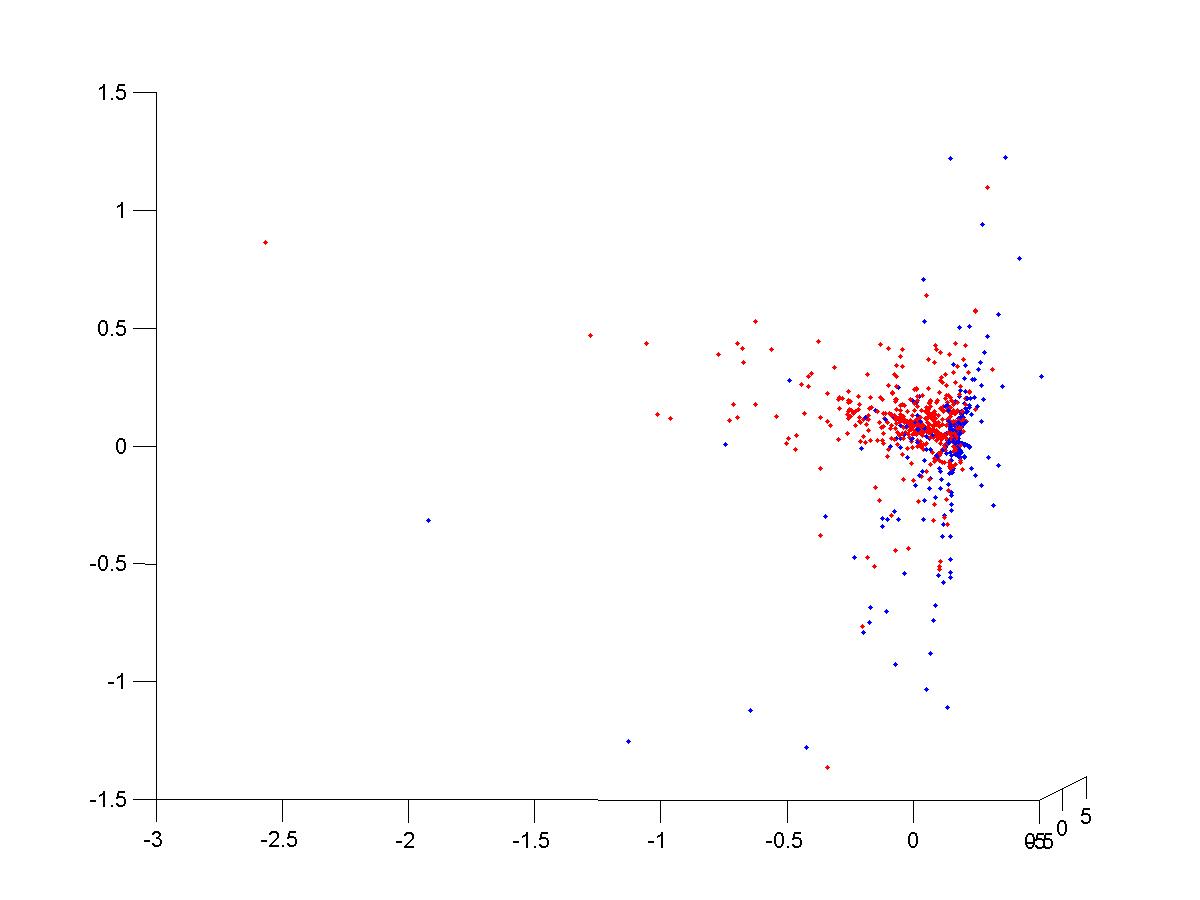}
\includegraphics[width=0.45\textwidth]{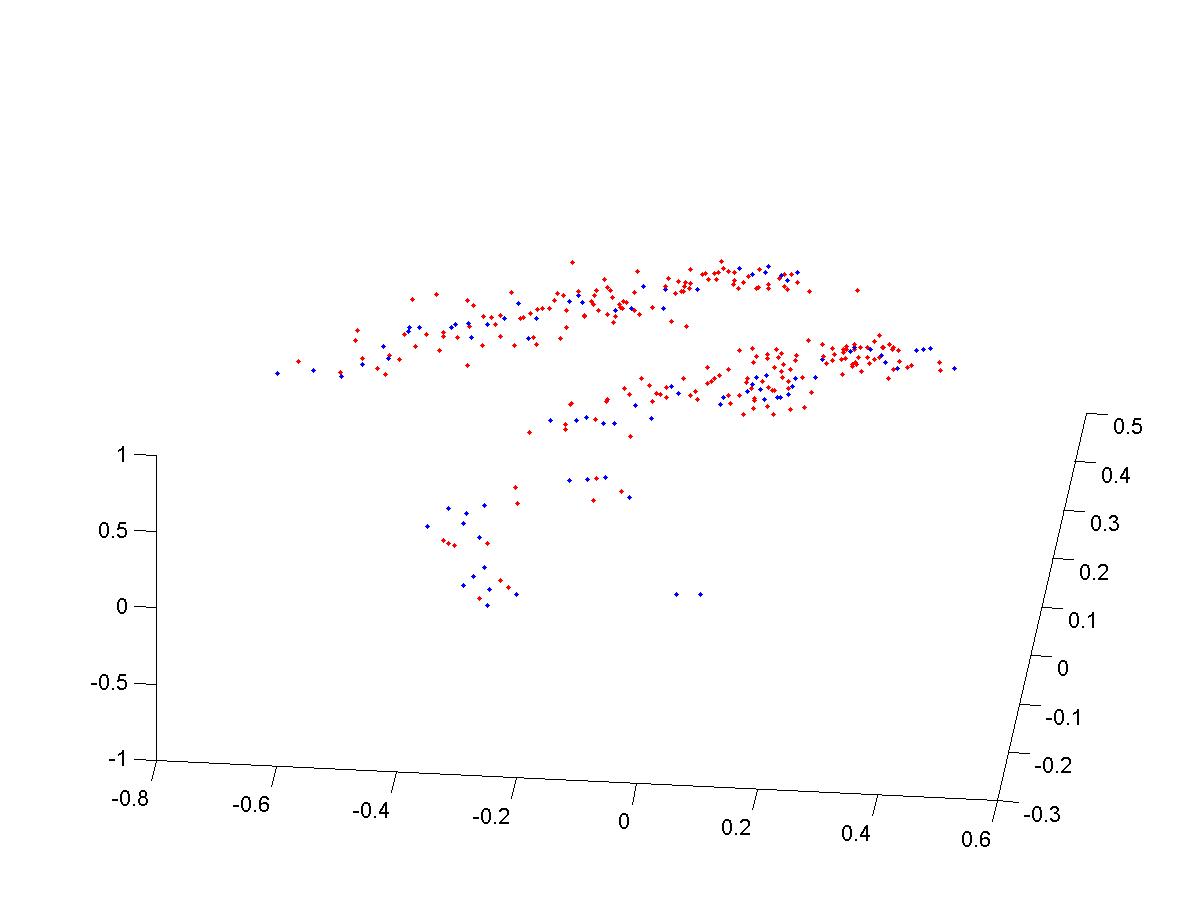}
\end{center}
\caption{SVD using adjectives, bag-of-words above, choices below}
\label{svdadjectives}
\end{figure}

\begin{figure}
\begin{center}
\includegraphics[width=0.45\textwidth]{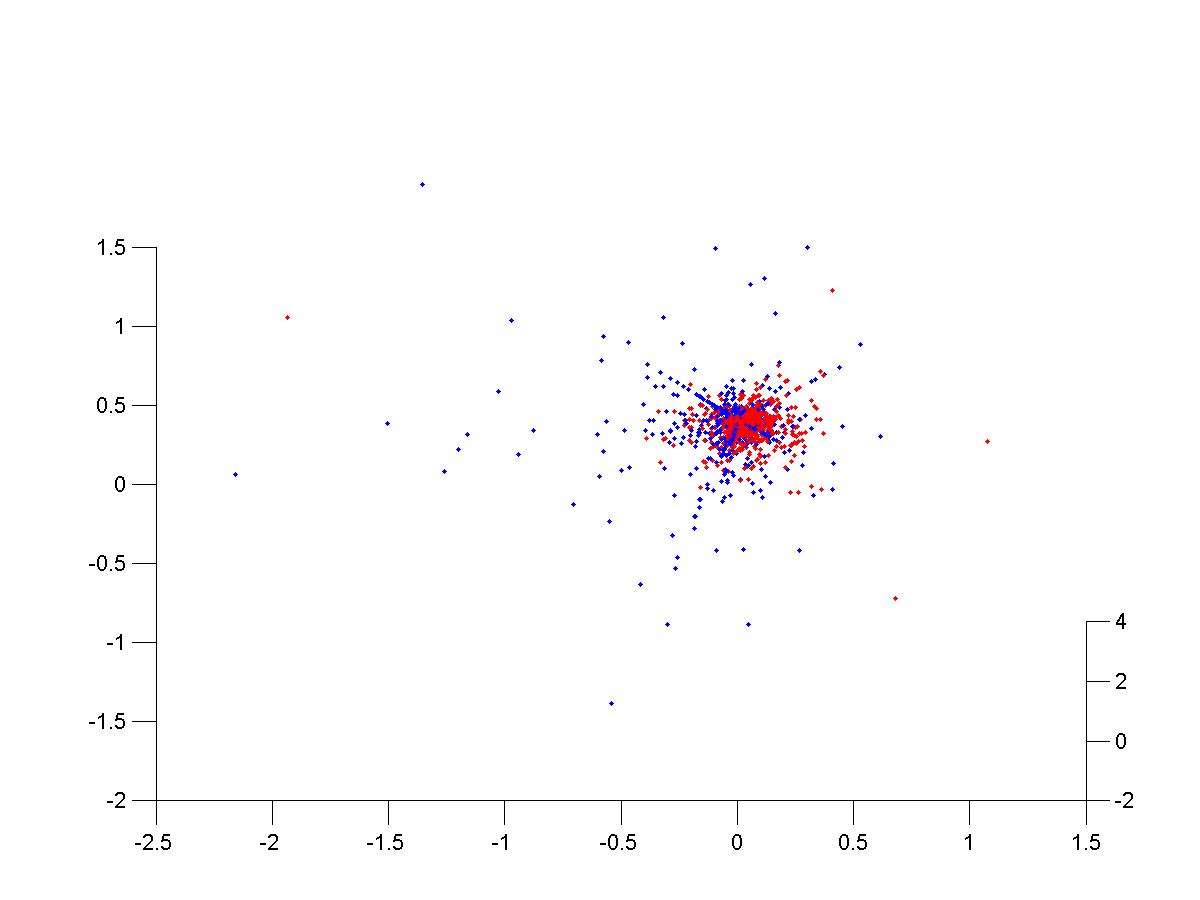}
\includegraphics[width=0.45\textwidth]{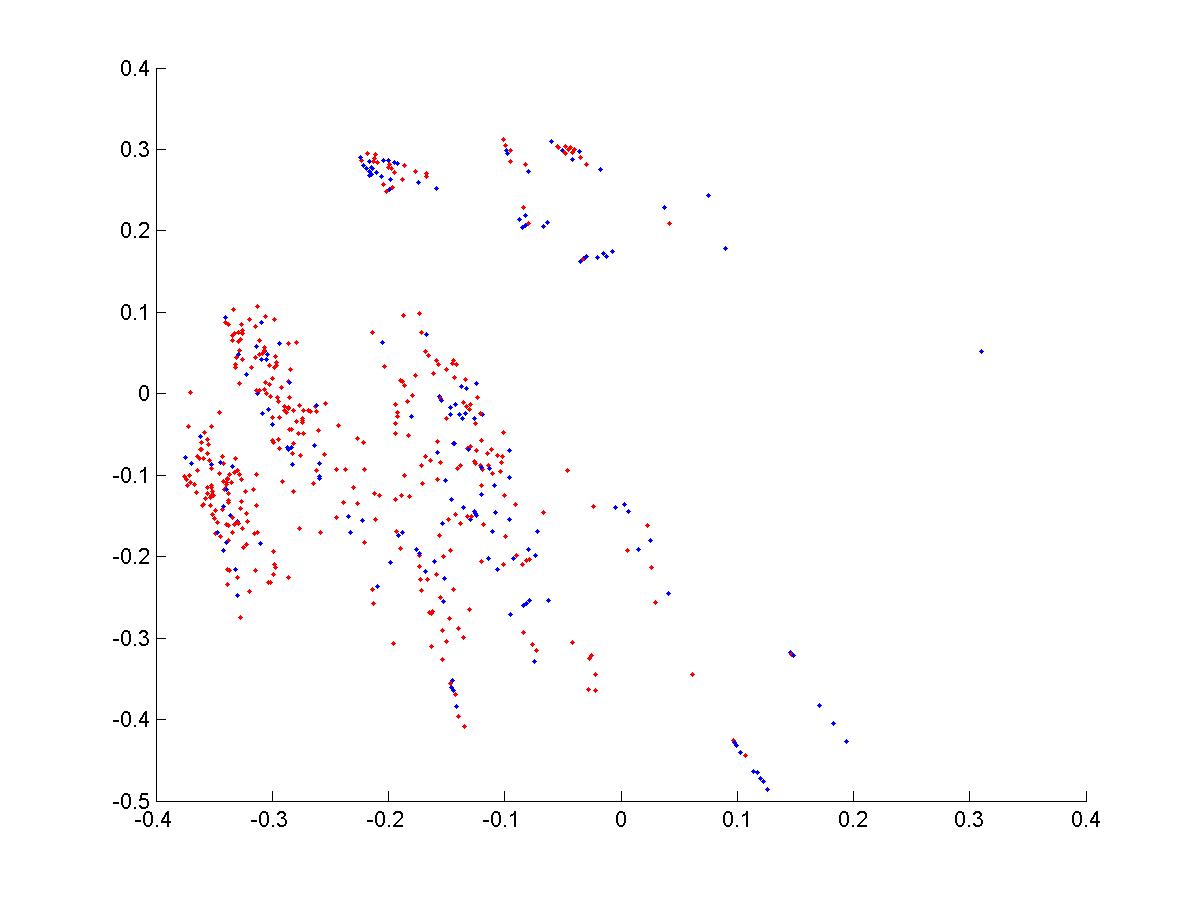}
\end{center}
\caption{SVD using verbs, bag-of-words above, choices below}
\label{svdverbs}
\end{figure}

We compute the singular value decomposition of the paragraph-word
matrix and the paragraph-choices matrix, both suitably normalized.
Plots show the resulting clustering of the paragraphs, with one
test novel's paragraphs in red and the other in blue.
In all of Figures \ref{svdpronouns}, \ref{svdaux},
\ref{svdadjectives}, and \ref{svdverbs} the clustering derived
from word frequencies is a single central cluster. In some of
them, there appears to be a separation between the two test documents,
but these are illusions caused by overlays of points.
In contrast, the clustering using choices shows strong clusters.
These correspond to paragraphs that resulted from similar
patterns of choices, that is to registers.

\subsection{Authorship prediction using word choices versus net choices}

We argued that systemic nets are useful for applications where properties
other than simple content are significant. To justify this claim we
predict authorship \emph{at the level of each individual paragraph}
for our two test novels. This is a difficult task because paragraphs
are so short; even humans would find it difficult to predict
authorship at this level, especially without access to the semantics
of the words used. 
Our goal is to show that the choice structure of the nets improves
performance over simple use of bags of words. There are, of course,
other ways to predict authorship, for example word n-grams, but
these are not directly comparable to systemic net approaches.

Again we use paragraph-word and paragraph-choice matrices
as our data, and 5-fold cross-validated support vector machines with
a radial basis kernel as the predictors. Results are shown for
each of the word sets in 
Tables~\ref{ppconfusionmatrix}, \ref{adverbsconfusionmatrix},
\ref{auxverbsconfusionmatrix}, \ref{posauxconfusionmatrix},
\ref{adjectivesconfusionmatrix} and \ref{verbsconfusionmatrix}.

\begin{table}
  \caption{Confusion matrices for personal pronouns; accuracy using words: 69.7\%, accuracy using choices: 75.3\%}
  \label{ppconfusionmatrix}
  \begin{tabular}{|c|c|c|c|c|}
  \cline{1-5}

  \multicolumn{1}{ |c| }{\multirow{2}{*}{Actual}} & \multicolumn{4}{ c| }{Predicted: words and choices} \\
  \cline{2-5}
  & \multicolumn{1}{|c}{RobCrusoe} & WutHeights & RobCrusoe & WutHeights \\ 
\cline{1-5}

  RobCrusoe &
  	  694 (48\%) & 33 (2\%) & 584 (40\%) & 143 (10\%)   \\ 
 \cline{1-5}
  	
  WutHeights &
  	  407 (28\%) & 320 (22\%) & 216 (15\%) & 511 (35\%)  \\ 
 \cline{1-5}
  \end{tabular}
\end{table}

\begin{table}
  \caption{Confusion matrices for adverbs; accuracy using words: 51.3\%, accuracy using choices: 63.4\%}
  \label{adverbsconfusionmatrix}
  \begin{tabular}{|c|c|c|c|c|}
  \cline{1-5}

  \multicolumn{1}{ |c| }{\multirow{2}{*}{Actual}} & \multicolumn{4}{ c| }{Predicted: words and choices} \\
  \cline{2-5}
  & \multicolumn{1}{|c}{RobCrusoe} & WutHeights & RobCrusoe & WutHeights \\ 
\cline{1-5}

  RobCrusoe &
  	  171 (12\%) & 556 (38\%) & 387 (27\%) & 340 (23\%)   \\ 
 \cline{1-5}
  	
  WutHeights &
  	  152 (10\%) & 575 (40\%) & 192 (13\%) & 535 (37\%)  \\ 
 \cline{1-5}
  \end{tabular}
\end{table}

\begin{table}
  \caption{Confusion matrices for auxiliary verbs; accuracy using words: 50.6\%, accuracy using choices: 72.0\%}
  \label{auxverbsconfusionmatrix}
  \begin{tabular}{|c|c|c|c|c|}
  \cline{1-5}

  \multicolumn{1}{ |c| }{\multirow{2}{*}{Actual}} & \multicolumn{4}{ c| }{Predicted: words and choices} \\
  \cline{2-5}
  & \multicolumn{1}{|c}{RobCrusoe} & WutHeights & RobCrusoe & WutHeights \\ 
\cline{1-5}

  RobCrusoe &
  	  435 (30\%) & 292 (20\%) & 553 (38\%) & 174 (12\%)   \\ 
 \cline{1-5}
  	
  WutHeights &
  	  426 (29\%) & 301 (21\%) & 233 (16\%) & 494 (34\%)  \\ 
 \cline{1-5}
  \end{tabular}
\end{table}

\begin{table}
  \caption{Confusion matrices for positive auxiliary verbs; accuracy using words: 51.4\%, accuracy using choices: 67.6\%}
  \label{posauxconfusionmatrix}
  \begin{tabular}{|c|c|c|c|c|}
  \cline{1-5}

  \multicolumn{1}{ |c| }{\multirow{2}{*}{Actual}} & \multicolumn{4}{ c| }{Predicted: words and choices} \\
  \cline{2-5}
  & \multicolumn{1}{|c}{RobCrusoe} & WutHeights & RobCrusoe & WutHeights \\ 
\cline{1-5}

  RobCrusoe &
  	  453 (30\%) & 292 (20\%) & 623 (43\%) & 104 (7\%)   \\ 
 \cline{1-5}
  	
  WutHeights &
  	  415 (29\%) & 312 (21\%) & 367 (25\%) & 360 (25\%)  \\ 
 \cline{1-5}
  \end{tabular}
\end{table}

\begin{table}
  \caption{Confusion matrices for adjectives; accuracy using words: 50.1\%, accuracy using choices: 70.8\%}
  \label{adjectivesconfusionmatrix}
  \begin{tabular}{|c|c|c|c|c|}
  \cline{1-5}

  \multicolumn{1}{ |c| }{\multirow{2}{*}{Actual}} & \multicolumn{4}{ c| }{Predicted: words and choices} \\
  \cline{2-5}
  & \multicolumn{1}{|c}{RobCrusoe} & WutHeights & RobCrusoe & WutHeights \\ 
\cline{1-5}

  RobCrusoe &
  	  295 (20\%) & 432 (30\%) & 490 (34\%) & 237 (16\%)   \\ 
 \cline{1-5}
  	
  WutHeights &
  	  294 (20\%) & 433 (30\%) & 187 (13\%) & 540 (37\%)  \\ 
 \cline{1-5}
  \end{tabular}
\end{table}

\begin{table}
  \caption{Confusion matrices for verbs; accuracy using words: 50.1\%, accuracy using choices: 67.5\%}
  \label{verbsconfusionmatrix}
  \begin{tabular}{|c|c|c|c|c|}
  \cline{1-5}

  \multicolumn{1}{ |c| }{\multirow{2}{*}{Actual}} & \multicolumn{4}{ c| }{Predicted: words and choices} \\
  \cline{2-5}
  & \multicolumn{1}{|c}{RobCrusoe} & WutHeights & RobCrusoe & WutHeights \\ 
\cline{1-5}

  RobCrusoe &
  	  297 (20\%) & 430 (30\%) & 476 (33\%) & 251 (17\%)   \\ 
 \cline{1-5}
  	
  WutHeights &
  	  296 (20\%) & 431 (30\%) & 221 (15\%) & 506 (35\%)  \\ 
 \cline{1-5}
  \end{tabular}
\end{table}

Across all of these word classes, authorship prediction based
on word use hovers close to chance; in contrast, authorship
prediction using systemic net choices range from accuracies of
around 65\% to 75\%, that is performance lifts of between 
15 and 20 percentage points over prediction from word choices.
Clearly, the structural information coded in the systemic nets
makes discrimination easier.

\subsection{Inferred nets versus randomly generated nets}

Tables~\ref{randompronoun} and \ref{randomadjective} compare the
authorship prediction performance of the inferred systemic net and
random networks constructed to have the same shape by dividing
the words hierarchically into nested subsets of the same sizes as
in the systemic net, but at random.

\begin{table}
     \caption{Personal pronouns: systemic network versus random nets}
   \label{randompronoun}
 \begin{tabular}{|c|c|c|c|c|c|}
     \cline{1-4}
      \hline
     Number of & \multicolumn{1}{ c| }{NNMF systemic network} & \multicolumn{3}{ c| }{Random nets} \\ \cline{2-5}

paragraphs & \multicolumn{1}{ |c| }{Accuracy} & {min} & mean
     & {max} \\ \cline{1-4}
   
     \hline
     \multicolumn{1}{ |c } {1} &
     \multicolumn{1}{ |c| } {75.3\% } & 69.3\% & 75.4\% & 82\%    \\ \cline{1-4}
   
     \hline
     \multicolumn{1}{ |c } {3} &
     \multicolumn{1}{ |c| } {84.1\% } & 68.1\% &72.1\% & 76.2\%    \\ \cline{1-4}
   
     \hline
     \multicolumn{1}{ |c } {6} &
     \multicolumn{1}{ |c| } {88.9\% } & 66.3\% & 70\% & 71.5\%    \\ \cline{1-4}
     \hline  
    \end{tabular}
\end{table}
  
\begin{table}
     \caption{Adjectives systemic network versus random nets}
     \label{randomadjective}

     \begin{tabular}{|c|c|c|c|c|c|}
      \cline{1-4}
      \hline
     Number of & \multicolumn{1}{ c| }{NNMF systemic network} & \multicolumn{3}{ c| }{Random nets} \\ \cline{2-5}

paragraphs & \multicolumn{1}{ |c| }{Accuracy} & {min} & mean
     & {max} \\ \cline{1-4}
   
      \hline
      \multicolumn{1}{ |c } {1} &
      \multicolumn{1}{ |c| } {70.8\% } & 70.2\% & 75.2\% & 79.4\%    \\ \cline{1-4}
    
      \hline
      \multicolumn{1}{ |c } {3} &
      \multicolumn{1}{ |c| } {72.3\% } & 66.9\% & 71.5\% & 73\%    \\ \cline{1-4}
    
      \hline
      \multicolumn{1}{ |c } {6} &
      \multicolumn{1}{ |c| } {74.8\% } & 64.5\% & 69.4\% & 72.3\%    \\ \cline{1-4}
      \hline
    
     \end{tabular}
  \end{table}

The performance of the random network is approximately the same
as the inferred network at the level of single paragraph prediction.
This is clearly a small sample size effect: choices that differentiate
authors well are also available in the random network by chance. 
However, as the number of paragraphs available to make
the prediction increases, the predictive performance of the systemic
net continues to improve while that of the random network remains flat.

\subsection{Combining systemic nets}

We have built our systemic nets starting from defined word sets. In
principle, a systemic net for all words could be inferred from a corpus.
However, such a net would represent, in a sense, the entire language
generation mechanism for English, so it is unlikely that it could be
reliably built, and would require an enormous corpus.

However, it is plausible that the systemic nets we have built could be
composed into larger ones, joining them together with an implied
conjunctive choice at the top level. We now investigate this possibility.

One way to tell if such a composition is meaningful is to attempt the
authorship prediction task using combined systemic nets. The results
are shown in Table~\ref{combined}. The combined nets show a lift of
a few percentage points over the best single net.

      \begin{table}
        \caption{Prediction accuracy using combined word sets,
best single systemic network, and combinations of systemic networks.}
        \label{combined}
      \begin{center}
       \begin{tabular}{| p{0.2\textwidth} | c | c | c |}
          \hline
  & words & best single & combined \\ \hline
Pronouns + adverbs & 69\% & 75.3\% & 77.4\% \\ \hline
Pronouns + adverbs + verbs & 73.1\% & 75.3\% & 80.2\% \\ \hline
Pronouns + adverbs + verbs + adjectives & 80.37\% & 75.3\% & 80.44\% \\ \hline
          \end{tabular}
      \end{center}
        \end{table}

These results hint, at least, that complex systemic nets can be built
by inferring nets from smaller sets of words, which can be done
independently and perhaps robustly; and then composing these
nets together to form larger ones. Some care is clearly needed: if
the choice created by composing two nets interacts with the choices
inside one or both of them, then the conjunctive composition may be misleading.
This property is known as selectional restriction, and is quite well
understood, so that it should be obvious when extra care is needed.
For example, composing a net for nouns and one for adjectives 
using a
conjunctive choice is unlikely to perform well because the choice of
a noun limits the choice of adjectives that ``match" it.

\section{Conclusions}

For large sets of documents, techniques based on bags of words have
been successful for tasks that have the flavor of information
retrieval, that is they depend only on the content of each
document. However, there are many other tasks where the significance
of each document depends not only on its content, but how it was
written (the mental states and abilities of the author), and
for what purpose (the social context). For these tasks, systemic
functional approaches have seemed attractive for some time; but their
application has been limited by the difficulty and expense of
constructing them. Here, we show that useful systemic nets can be
inferred inductively from example corpora; and that the resulting
nets, although not matching standard linguistic intuitions, are
nevertheless useful for both clustering and prediction.

\bibliographystyle{plain}

\bibliography{../../biblio/dm}
\end{document}